\documentclass[graybox]{svmult}

\usepackage{type1cm}        
\usepackage{makeidx}         
\usepackage{graphicx}        
\usepackage{multicol}        
\usepackage[bottom]{footmisc}

\newcommand{\pkg}[1]{{\normalfont\fontseries{b}\selectfont #1}}
\let\proglang=\textsf

\usepackage{makecell}
\usepackage{newtxtext}       %
\usepackage[varvw]{newtxmath}       
\usepackage{geometry}
\makeindex             
\usepackage{natbib}
\bibpunct[, ]{(}{)}{;}{a}{}{,}
\usepackage{hyperref}
\def\equationautorefname~#1\null{Equation~(#1)\null}
\hypersetup{
  colorlinks=true,
  linkcolor=blue,
  citecolor=blue,
  urlcolor=blue}

\begin{document}

\title*{Forecasting large collections of time series: feature-based methods}
\author{Li Li, Feng Li and Yanfei Kang}
\institute{
Li Li \at School of Economics and Management, University of Science \& Technology Beijing, Beijing, China, ORCID: \texttt{0000-0002-7922-1281}, \email{li.li@ustb.edu.cn}.
\and Feng Li \at School of Statistics and Mathematics, Central
University of Finance and Economics, Beijing, China, ORCID: \texttt{0000-0002-4248-9778}, \email{feng.li@cufe.edu.cn}.
\and Yanfei Kang. Corresponding author.  \at School of Economics and Management, Beihang University, Beijing, China, ORCID: \texttt{0000-0001-8769-6650}, \email{yanfeikang@buaa.edu.cn}.}
%
%
\maketitle

\abstract{
In economics and many other forecasting domains, the real world problems are too complex for a single model that assumes a specific data generation process. The forecasting performance of different methods changes depending on the nature of the time series. When forecasting large collections of time series, two lines of approaches have been developed using time series features, namely feature-based model selection and feature-based model combination. This chapter discusses the state-of-the-art feature-based methods, with reference to open-source software implementations.
}

\section{Introduction}

\begin{figure}
\label{fig:framework}
\centering
\includegraphics[width=\textwidth]{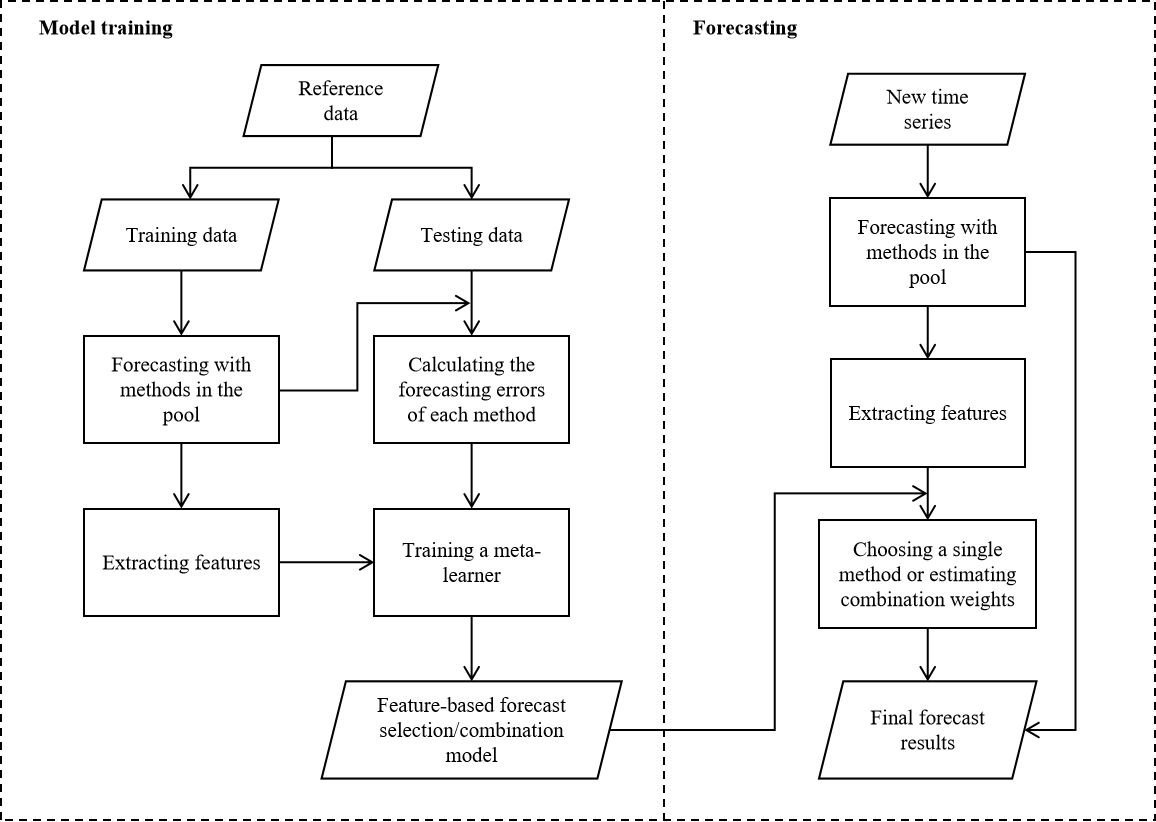}
\caption{The framework of feature-based methods for forecasting.}
\end{figure}

When the vast collection of time series is a common case in economic and other forecasting domains, AI empowers rapid decision-making. The No-Free-Lunch theorem \citep{wolpert1997no} tells that no single model always performs the best for all time series. When forecasting large collections of time series, instead of choosing one model for all the data, features can be used to obtain the most appropriate model or the optimal combination of candidate models, per series. With the advancement of AI, more recent literature uses meta-learning to describe the process of automatically acquiring knowledge for forecast model selection/combination, which can be summarized in a general framework in Figure~\ref{fig:framework}. Specifically, the framework is divided into two phases.
In the model-training phase, we need a set of reference data that splits into training and testing periods. Then a meta-learner is used to learn the relationship between individual forecasts from the pool and the forecasting performance measured by an error metric. Finally, a forecast selection/combination algorithm is trained by minimizing the total forecasting loss. Once the model has been trained, the best forecast method or the combination weights can be decided for any target series given the features.

The idea of meta-learning could date back to the 1970s, when \cite{rice1976algorithm} proposed a groundbreaking framework for algorithm selection problems. \cite{collopy1992rule} developed 99 rules based on 18 features. \cite{shah1997model} used several features to classify time series and applied discriminant analysis to predict the best-performing model. The term ``meta-learning''  emerged with machine-learning literature. \cite{prudencio2004meta} presented an original work that used  ``meta-learning'' in model selection for time-series forecasting. They investigated two meta-learning approaches based on two case studies. \cite{wang2009rule} focused rule induction for selecting a forecasting method by understanding the nature of historical data. \cite{lemke2010meta} compared different meta-learning approaches to investigate which model worked best in which situation.  More recently, \citep{kuck2016meta} proposed a meta-learning framework based on neural networks for forecast selection. Further evidence in favor of meta-learning is given in \cite{talagala2023metalearning} and \cite{talagala2022fformpp}.

Using meta-learning to obtain weights for forecast combinations has received increased attention lately. The feature-based forecast model averaging (FFORMA) framework proposed by \cite{montero2020fforma} employed 42 features to estimate the optimal combination weights of nine forecasting methods based on extreme gradient boosting (XGBoost, \cite{chen2016xgboost}). \cite{kang2020gratis} used 26 features to predict the performances of nine forecasting methods with synthetic time series data, and obtained the combination weights by a tailored softmax function of the predicted forecasting errors.  \cite{li2020forecasting}, \cite{kang2022forecast}, \cite{wang2022uncertainty}, and \cite{talagala2022fformpp} extended the framework of FFORMA based on new types of features and a variety of meta-learners and real world applications. The representative research in feature-based forecasting is summarized in Table \ref{tab}.

\begin{table}[htbp]
  \centering
  \caption{Overview of literature on feature-based forecasting}
    \begin{tabular}{p{4cm} p{2cm} p{3cm} p{2cm} }
    \hline
   Authors (Year) &  Type  &  Meta-learning algorithm &{\makecell[c]{Number of \\ features used}} \\
    \hline
    \cite{collopy1992rule} &  Combination  &  Rule-based induction & {\makecell[c] {18}}\\
    \cite{shah1997model} & Selection & Discriminant analysis & {\makecell[c] {26}}\\
\cite{prudencio2004meta} & Selection & C4.5 decision tree, Neural network classifier & {\makecell[c] {14, 16}}\\
\cite{wang2009rule} & Selection & C4.5 decision tree, SOM clustering & {\makecell[c] {9}}\\
\cite{lemke2010meta} & Selection & Feed-forward neural
network, decision tree, support vector machine & {\makecell[c] {24}}\\
\cite{petropoulos2014horses} & Selection & Neural network & {\makecell[c] {9}}\\
\cite{kuck2016meta} & Selection & Neural network & {\makecell[c] {127}}\\
\cite{talagala2023metalearning} & Selection & Random forest & {\makecell[c] {33}}\\
\cite{kang2020gratis} & Combination & Nonlinear regression & {\makecell[c] {26}}\\
\cite{montero2020fforma} & Combination & XGBoost & {\makecell[c] {42}}\\
\cite{kang2022forecast} & Combination & XGBoost &
Depend on the number of forecasting models\\
\cite{li2020forecasting} & Combination & XGBoost &
Depend on time series imaging \\
\cite{li2022feature} & Combination & XGBoost & {\makecell[c] {9}}\\
\cite{wang2022uncertainty} & Combination &
{\makecell[l]{Generalized additive \\ models (GAMs)}}  & {\makecell[c] {43}} \\
\cite{talagala2022fformpp} & Selection & Bayesian multivariate surface regression & {\makecell[c] {37}}\\
    \hline
    \end{tabular}%
  \label{tab}%
\end{table}%

In practice, a typical feature-based forecasting method requires the forecaster to make the following decisions:
1) What training data to use to train the meta-learner?
2) What features to use to describe the time series characteristics?
3) What forecasts to include to formulate the forecast pool?
4) How to use meta-learners in applications?
As such, around the above questions, the remainder of this chapter includes the following sections.

\begin{itemize}
\item \textbf{Data generation}. The performance of any time series mining algorithm, including forecasting, depends on the diversity of the training data. In the era of big data, many industries have ample and diverse historical data that can be used to train forecasting models. However, there are still cases such as the lack of historical data, or it would be very difficult or expensive to obtain data. To solve these problems, we introduce some methods for time series generation.

\item \textbf{Feature extraction}. This section introduces a series of time series features widely used in forecasting and ways to automatically extract time series features without the need for expert knowledge and human interaction. The methods for the automation of feature extraction include time series imaging, calculation of forecast diversity and automatic feature selection from thousands of features.

\item \textbf{Forecast trimming}. This section discusses methods to decide which forecasts should be included especially for combinations. The gains from forecast combinations highly depend on the quality of the pool of forecasts to be combined and the estimation of combination weights. There are three main factors to influence the quality of the forecasting pool, which are accuracy, robustness (measured by the variance of the accuracy), and diversity (independent information contained in the component forecasts) of the individual forecasts. We review some classical methods and provide a way to address the accuracy, robustness, and diversity simultaneously.

\item \textbf{Some practical forecasting issues}. This section discusses some practical issues that arise in feature-based forecasting. For example, intermittent demand with several periods of zero demand is ubiquitous in practice. In addition, uncertainty estimation can indicate risk information, which is useful in actual decisions. We provide some feature-based solutions for the two aforementioned issues. The choices of meta-learners, as well as loss functions, are also discussed.
\end{itemize}

\section{Data Generation}
\label{sec:1}
The performance of any time series forecasting method depends on the diversity of the training data, so that the evaluation of the method can be generalized to a wide range of future data. Therefore, there is a need for comprehensive time series benchmarks. Although some attempts have been made in certain time series tasks, such as the M4 dataset for a recent time series forecasting competition \citep{makridakis2020m4}, the time series area lacks diverse and controllable benchmarking data for forecasting evaluation. This section reviews some recent work for time series generation.  Section \ref{sec:generation} introduces a series of methods to generate diverse time series and mainly focuses on Gaussian mixture autoregressive (MAR) models.  Section \ref{sec:generation_controllable} reviews some advanced methods that can generate time series with controllable characteristics.

\subsection{Diverse time series generation}
\label{sec:generation}

In recent years, research has shown great potential to use generated data for algorithm learning under certain application domains. Such examples can be found in evaluating statistical methods for temporal outbreak detection \citep{lotze2009does, bedubourg2017evaluation}, examining neural network forecasting \citep{zhang2001simulation}, and so on. In previous studies of time series generation, some scholars focused on the shapes of given time series, or on some predefined types of time series. \cite{vinod2009maximum} used the maximum entropy bootstrap to generate ensembles for time series inference by extending the traditional iid bootstrap to nonstationary and dependent data. \cite{bagnall2017simulated}  simulated time series from different shape settings by placing one or more shapes on a white noise series. The simulators are designed for different feature spaces to evaluate classification algorithms. But it is impossible to create simulated time series covering the possible space of all time series, which limits the reproducibility and applicability of the tested methodology.

\cite{kang2020gratis} proposed an efficient and general approach of GeneRAting TIme Series (gratis) with diverse and controllable characteristics, named GRATIS, based on MAR models to generate a wide range of non-Gaussian and nonlinear time series. Mixture transition distribution models were first developed by \cite{li2010flexible} to capture non-Gaussian and nonlinear features, and generalized to MAR models later \citep{wong2000mixture}.

MAR models consist of multiple stationary or nonstationary autoregressive components. A \textit{K}-component MAR model can be defined as:
\begin{equation}
\label{eq:mar}
F\left(x_t \mid \mathbb{F}_{-t}\right)=\sum_{k=1}^K \alpha_k \Phi\left(\frac{x_t-\phi_{k 0}-\phi_{k 1} x_{t-1}-\cdots-\phi_{k p_k} x_{t-p_k}}{\sigma_k}\right),
\end{equation}
where $F\left(x_t \mid \mathbb{F}_{-t}\right)$ is the conditional cumulative distribution
of $x_t$ given the past information $\mathbb{F}_{-t} \subseteq\left\{x_{t-1}, \ldots, x_{t-p_k}\right\}$, $\Phi(\cdot)$ is the cumulative distribution function of the standard normal distribution, $x_t-\phi_{k 0}-\phi_{k 1} x_{t-1}-\cdots-\phi_{k p_k} x_{t-p_k}$ is
the autoregressive term in each mixing component, $\sigma_k>0$ is the standard error, $\sum_{k=1}^K \alpha_k=1$ and $\alpha_k>0$ for \textit{k} = 1, 2, …, \textit{K}. Denoted as $\operatorname{MAR}\left(K ; p_1, p_2, \ldots, p_k\right)$, it is actually finite mixtures of \textit{K} Gaussian AR models.

A significant difference in \cite{kang2020gratis}'s data generation process compared to typical simulation processes used in the literature is that the authors used distributions instead of fixed values for the parameters. This allows for the generation of diverse time series instances.  The parameter settings are analogous to noninformative priors in the Bayesian contexts, that is, the diversity of the generated time series should not rely on the parameter settings.

So far, we have focused on time series with only one seasonal pattern. However, many time series exhibit multiple seasonal patterns of different lengths, especially those series observed at a high frequency (such as daily or hourly data). Simulation of multi-seasonal time series involves the weighted aggregation of simulated time series with the corresponding frequencies. A simulated multiseasonal time series $x_t$ with \textit{M} seasonal patterns can be written as $x_t=\sum_{m=1}^M \omega_m x_{F_m, t}$, where $m=1,2, \ldots, M$, $x_{F_m, t}$ is the \textit{m}th simulated time series with frequency $F_m$, and weight $\omega_m$ satisfies $\sum_{m=1}^M \omega_m=1$ and $0<\omega_m <1$. The weights can be obtained by $\omega_m=\frac{\gamma_m}{\sum_{r=1}^M \gamma_r}$, where $\gamma_m \sim U(0,1)$.

The critical merits of MAR models for nonlinear time series modeling are \citep{wong2000mixture,li2010flexible}: (a) MAR models can capture extensive time series features in principle based on a sufficiently diverse parameters space and a finite number of components, (b) mixtures of stationary and nonstationary components can yield both stationary and nonstationary process with MAR, (c) the conditional distributions of time series change with time, which allows for evolution with historical information, (d) MAR models can handle complicated univariate and multivariate time series with different values of frequencies and seasonality, and (e) MAR models can also capture features such as multimodality, heavy tails and heteroskedasticity.

\subsection{Time series generation with target features}
\label{sec:generation_controllable}
In time series analysis, researchers with a particular focus may be only interested in a specific area of the feature space or a subset of features, for example, heteroskedasticity and volatility in financial time series, trend and entropy in microeconomic time series, or peaks and spikes in energy time series. Practitioners may also want to mimic more time series from their limited collection of real data, such as sales records for new products and health metrics for rare diseases. Therefore, the efficient generation of time series with target features of interest is another crucial problem to address.

\cite{kang2017visualising} used a genetic algorithm (GA) to generate new time series to fill in any gaps in a two-dimensional instance space. \cite{kegel2017generating} applied STL method to estimate the trend and seasonal component of a time series, which are modified using multiplicative factors to generate new time series.
\cite{kang2017visualising}'s evolutionary algorithm is quite general, but computationally slow, whereas the STL-based algorithm from \cite{kegel2017generating}  is much faster but can only generate series that are additive in trend and seasonality.
\cite{talagala2023metalearning} augmented the set of observed time series by simulating new time series similar to those to form a larger dataset for training the model-selection classifier. The simulated series rely on the assumed data generating processes (DGPs), which are exponential smoothing models and ARIMA models.

\cite{kang2020gratis} used a GA to tune the MAR model parameters until the distance between the target feature vector and the feature vector of a sample of time series simulated from the MAR is close to zero. The parameters for the MAR model, the underlying DGP, can be represented as a vector $\Theta=\left\{\alpha_k, \phi_i\right\}$, for $k=1, \ldots, K$, $i=k 0, \ldots, k p_k$ in Equation (\ref{eq:mar}). The authors developed an R package \texttt{gratis} for the time series generation which is available from CRAN. The R package \texttt{gratis} provides efficient algorithms for generating time series with diverse and controllable characteristics.

\section{Feature Extraction}
\label{sec:2}
A feature can be any function computed from a time series. Examples include a simple mean, the parameter of a fitted model, or some statistic intended to highlight an attribute of the data. A unique “best” feature representation of a time series does not exist. What features are used depends on both the nature of the time series being analyzed and the purpose of the analysis. For example, consider the mean as a simple time series feature. If some time series contain unit roots, then the mean is not a meaningful feature without additional constraints on the initial values of the time series. Even if the series are all stationary, if the purpose of our analysis is to identify the best forecasting method, then the mean is probably of no value. This example shows that it is difficult to formulate general desirable properties of features without knowledge of both the time series properties and the required analysis. Section \ref{features} reviews a series of features that have been found helpful in time series exploration.

Recently, a series of automated approaches to extract time series features have been put forward, which do not need experts select features manually and require limited human interaction. Section \ref{sec:imaging} introduced a new tool of time series imaging proposed by \cite{li2020forecasting}. \cite{li2020forecasting} transformed time series into recurrence plots, from which local features can be extracted using computer vision algorithms. The study reviewed above depends on using each time series's historical, observed data. The estimation of the features might not be feasible or robust in the case of a limited number of available past observations. \cite{kang2022forecast} proposed to focus on the produced forecasts to extract features, which is discussed in Section \ref{sec:diversity}.  When the chosen features involve large numbers, estimating them might increase the computational time required. \cite{theodorou2022exploring} provided a methodological approach to select from thousands of features, which is discussed in Section \ref{sec:selection}.

\subsection{Time series features}
\label{features}
We use statistical functions to compute the features of that time series, such as the mean, minimum or maximum. For example, all the autocorrelation values of a series can be considered features of that series. We can also summarise the autocorrelation values to produce new features. The sum of the first ten squared autocorrelation coefficients is a useful summary of how much autocorrelation there is in a series. We can also compute the autocorrelation values of the changes in the series between periods. One could create a new time series consisting of the differences between consecutive observations. Occasionally it is useful to apply the same differencing operation again, so one could compute the differences of the differences. Another related approach is to compute seasonal differences of a series.  The autocorrelations of these differenced series may provide useful information. In addition, the STL (a seasonal-trend decomposition) is the basis for several features. A time series decomposition can be used to measure the strength of trend and seasonality in a time series.

Based on the literature review, time series features vary from tens to thousands, and choosing different sets of features will inevitably result in different forecasts and varied performance. Therefore, choosing and weighting time series features is a common challenge in feature-based forecasting. Several packages, including functions for computing features and statistics from time series, have been developed and widely used in the forecasting community. For example, the \pkg{tsfeatures}  \citep{Hyndman2019tsfeatures} and \pkg{feasts} \citep{Mitchell2022feasts} packages for \proglang{R} include statistics that are computed on the first and second-order differences of the raw series, account for seasonality, and exploit the outputs of popular time series decomposition methods, among others; the \pkg{tsfresh} \citep{christ2018time} package for \proglang{Python} involves thousands of features, such as basic time series statistics, correlation measures, entropy estimations, and coefficients of standard time series forecasting and analysis methods; the \pkg{catch22} \citep{lubba2019catch22} includes 22 time-series features selected from the over 7000 features in \pkg{hctsa} \citep{fulcher2013highly}. Several feature-based forecast combination models used the features in \pkg{tsfeatures} as the basic pool of features \citep{kang2020gratis,montero2020fforma,wang2022uncertainty,li2022bayesian,talagala2022fformpp}.

\subsection{Automation of feature extraction}

\subsubsection{Time series imaging}
\label{sec:imaging}

\begin{figure}
\centering
\includegraphics[width=\textwidth]{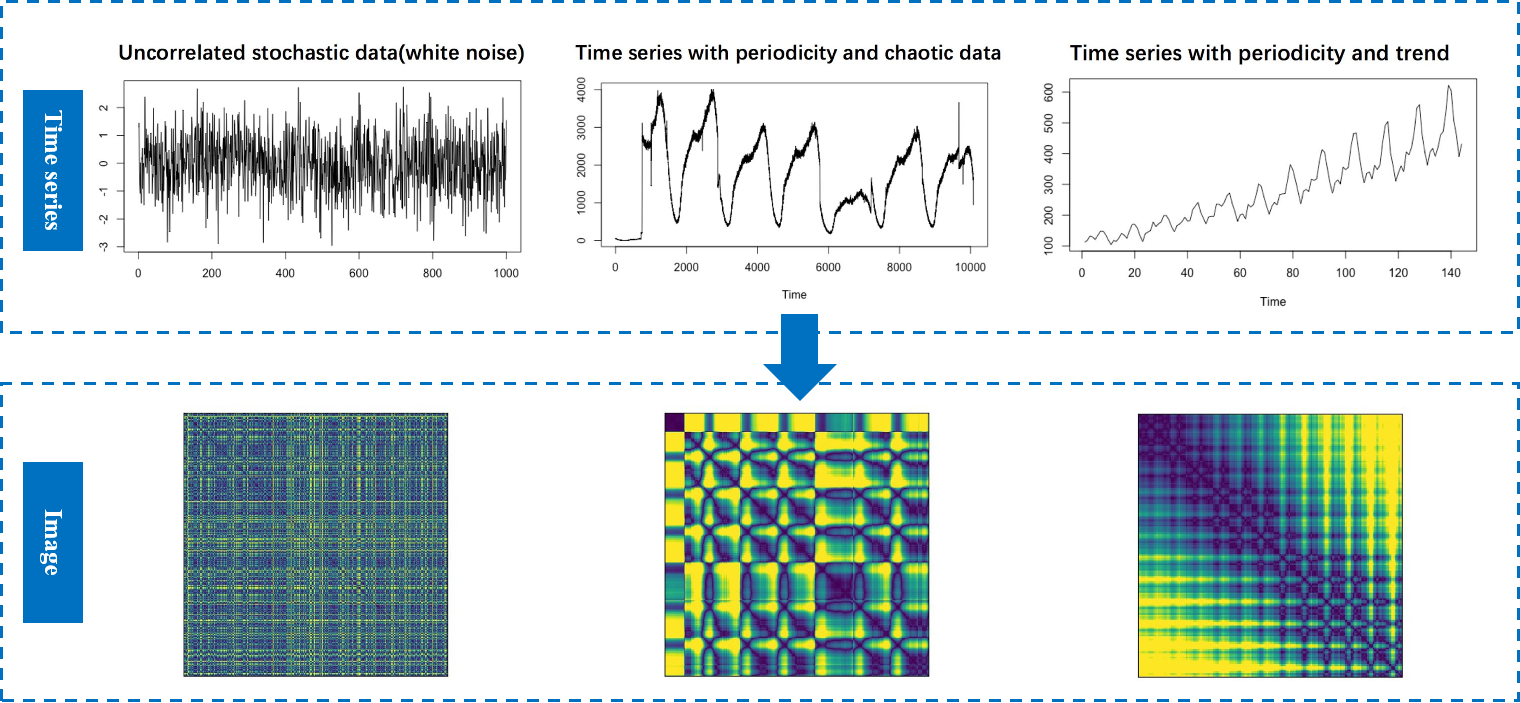}
\caption{Typical examples of recurrence plots (top row) for time series data with different patterns (bottom row): uncorrelated stochastic data, i.e., white noise (left), a time series with periodicity and chaos (middle), and a time series with periodicity and trend (right).}
\label{fig:imaging}
\end{figure}

\cite{li2020forecasting} introduced an automated approach to extract time series features based on a time series imaging process. In the first step, \cite{li2020forecasting} encoded the time series into images using recurrence plots. In the second step, image processing techniques extract time series features from images.  \cite{li2020forecasting} considered two different image feature extraction approaches: a spatial bag-of-features (SBoF) model and convolutional neural networks (CNNs). Recurrence plots (RPs) provide a way to visualize the periodic nature of a trajectory through a phase space and can contain all relevant dynamical information in the time series. Figure \ref{fig:imaging} shows three typical recurrence plots. They reveal different patterns of recurrence plots for time series with randomness, periodicity, chaos, and trend.

The SBoF model used in \cite{li2020forecasting} for time series feature extraction using consists of three steps: (i) detect key points with the scale-invariant feature transform (SIFT) algorithm and find basic descriptors with k-means; (ii) generate the representation based on the locality-constrained linear coding (LLC) method; and (iii) extract spatial information by spatial pyramid matching (SPM) and pooling. For the details in each step, see \cite{li2020forecasting}.

An alternative to SBoF for image feature extraction is to use a deep CNN, which has achieved great breakthroughs in image processing. In this task, the deep network is trained on the large ImageNet dataset, and the pre-trained network is publicly available. For the CNN model, the closer the layer is to the first layer, the more general features can be extracted; the closer the layer is to the back layer, the more specific features for classification tasks can be extracted. To extract time series features, the parameters of all the previous layers are fixed except for the last fully connected layer, and the last layer is fine-tuned. With the trained model, the final representation of the time series can be obtained. The code to implement the above algorithm is available at \url{https://github.com/lixixibj/forecasting-with-time-series-imaging}.

\subsubsection{Forecast diversity}
\label{sec:diversity}

The task of selecting appropriate features to build forecasting models is often challenging. Even if there was an acceptable way to define the features, existing features are estimated based on historical patterns, which are likely to change in the future. Other times, the estimation of the features is infeasible due to limited historical data. To solve these problems, \cite{kang2022forecast} suggested a change of focus from the historical data to the produced forecasts to extract features. \cite{kang2022forecast} used out-of-sample forecasts to obtain weights for forecast combinations by amplifying the diversity of combined methods.

For a given time series $\left\{y_t, t=1,2, \ldots, T\right\}$, the $h$th step forecast produced by the $i$th individual method is denoted as $f_{ih}$ , where $i = 1 , 2 , \ldots, M$ and $h = 1 , 2 , \ldots, H$. Furthermore, $M$ and $H$ are the number of methods in the forecast pools and the forecast horizon, respectively. Let $f_{ch}$ be the $h$th step combined forecast given by
$\sum_{i=1}^M w_i f_{ih}$, where $w_i$ is the combination weight for the $i$th method. The overall mean squared error of a weighted forecast combination model ${MSE}_{comb}$ over the whole forecast horizon $H$ can be written as follows.
\begin{equation}
\label{eq:mse}
\begin{aligned}
\operatorname{MSE}_{\text {comb }} & =\frac{1}{H} \sum_{i=1}^H\left(\sum_{i=1}^M w_i f_{i h}-y_{T+h}\right)^2 \\
& =\frac{1}{H} \sum_{i=1}^H\left[\sum_{i=1}^M w_i\left(f_{i h}-y_{T+h}\right)^2-\sum_{i=1}^M w_i\left(f_{i h}-f_{c h}\right)^2\right] \\
& =\frac{1}{H} \sum_{i=1}^H\left[\sum_{i=1}^M w_i\left(f_{i h}-y_{T+h}\right)^2-\sum_{i=1}^{M-1} \sum_{j>i}^M w_i w_j\left(f_{i h}-f_{j h}\right)^2\right]\\
& =\sum_{i=1}^M w_i M S E_i-\sum_{i=1}^{M-1} \sum_{j>i}^M w_i w_j D i v_{i, j},
\end{aligned}
\end{equation}
where ${MSE}_i$ represents the mean squared error for the $i$th method. ${Div}_{i,j}$ denotes the degree of diversity between the $i$th and $j$th method in the forecast method pool.

Equation (\ref{eq:mse}) says that the mean squared error of the combined forecast is guaranteed to be less than or equal to the weighted mean squared error of the individual forecasts. The second term in the last line of Equation (\ref{eq:mse}) tells us how diverse the individual forecasts are. The more diversity existing in the forecast method pool leads to overall better forecasting accuracy.

\cite{kang2022forecast} proposed to use the pairwise diversity measures as a set of features for each time series. To make the diversity comparable between time series with different scales, the scaled diversity can be defined as:
\begin{equation}
\label{eq:sDiv}
s{Div}_{i, j}=\frac{\sum_{h=1}^H\left(f_{i h}-f_{j h}\right)^2}{\sum_{i=1}^{M-1} \sum_{j=i+1}^M\left[\sum_{h=1}^H\left(f_{i h}-f_{j h}\right)^2\right]} .
\end{equation}

 Given a time series data set with $N$ time series and a forecasting pool containing $M$ methods,  $H$-step point forecasts are produced by the methods. Then we can get an $M \times H$ matrix for each time series to forecast. Based on Equation (\ref{eq:sDiv}), the pairwise forecast diversity among the $M$ methods can be calculated. Thus, for each time series, we get an $M \times M$ symmetric matrix, which can be transformed into a feature vector with  $M(M-1) / 2$  pairwise diversity measures.

The merits of using diversity for forecast combinations are twofold. First, the process of extracting diversity is straightforward and interpretable. The algorithm for measuring the diversity between different methods involves a simple calculation, and hence, it can reduce the computational complexity when extracting features. Secondly, although traditional feature extraction methods usually depend on the manual choice of an appropriate set of features, our approach can be applied automatically without needing expert knowledge and human interaction.

\subsubsection{Automatic feature selection}
\label{sec:selection}
A large number of features significantly increase the dimensionality and complexity. Therefore, a feature selection procedure is necessary. \cite{theodorou2022exploring} used three steps to capture the key features of the M5 competition data, including statistical pre-filtering, performance evaluation, and redundancy minimization.

The statistical pre-filtering step aims to remove the non-significant features rather than choosing the most meaningful ones. It involves the z-score standardization of the features and eliminating those that display the same or similar values across different series. The features that can effectively differentiate distinct classes in the data set in a statistically significant manner are selected.

To evaluate the quality of the extracted features, \cite{theodorou2022exploring} employed the Regressional ReliefF (RReliefF) algorithm \citep{robnik2003theoretical}, which is an extension of the ReliefF algorithm for regression problems. ReliefF is a robust filtering method used to select features in multi-class classification problems, with the basic idea of identifying feature differences between nearest instance pairs. Specifically, ReliefF calculates a score for each feature and performs feature selection accordingly. \cite{li2022bayesian} constructed a multi-class classification problem by labeling the model that performs the best and employed the ReliefF algorithm to rank the features. Then some of the top features are selected for \cite{li2022bayesian}'s Bayesian forecast combination framework. An R package \pkg{febama} is developed for this purpose at \url{https://github.com/lily940703/febama}.

\cite{theodorou2022exploring} employed hierarchical clustering to reduce the redundancy
in the obtained top-performing features using the Pearson correlation distance (cosine distance) with complete linkage at a threshold of 0.2. The process makes the pairwise correlation coefficients of the features in the same cluster larger than 0.8, thus forming clusters of similarly performing features. The vectors generated from the previous step (performance evaluation) are the input to the clustering method. In each cluster, the feature with the largest mean quality score is selected.

After completing the above feature selection procedure, \cite{theodorou2022exploring} obtained a set of 42 features, including information about the coefficients of
the discrete Fourier transform, the variance and distribution of
the data, the entropy and linear trend of the series, the statistics of popular tests for time series analysis, and so on. Although some of the selected features are challenging to interpret, the method of automatic feature selection is more flexible and generic compared with alternative methods that arbitrarily use a limited set of features.

\section{Forecast Trimming for Combination}
\label{sec:3}
Forecast combination is widely used as a preferred strategy over forecast selection due to its ability to reduce the uncertainty of identifying a single “best” forecast. Nonetheless, sophisticated combinations are often empirically defeated by simple averaging, which is commonly attributed to the weight estimation error. The issue becomes more serious when dealing with a forecast pool containing a large number of individual forecasts. Therefore, forecast trimming algorithms are indispensable to identify an optimal subset from the original forecast pool for forecast combination tasks.

When determining which forecasts should be combined, it is crucial to look at the characteristics of the available alternative forecasts, among which robustness, accuracy, and diversity are the
most frequently emphasized in the literature \citep{budescu2015identifying,thomson2019combining,atiya2020does,lichtendahl2020some}. Section \ref{sec:Accuracy} discusses the three factors and reviews some classical methods for forecast trimming. \cite{wang2022another}  took into account the robustness, accuracy and diversity of the forecast pool simultaneously and proposed a novel algorithm, which is discussed in Section \ref{sec:trimming}.

\subsection{Accuracy, robustness and diversity}
\label{sec:Accuracy}
\textbf{Accuracy.} Forecast combinations base their performance on the mean level of the accuracy of the individual forecasts to be combined. Including a poorly performing forecast in a combination is likely to decrease the accuracy of the combined forecast. Therefore, it makes intuitive sense to eliminate the worst performers from the forecast pool based on some performance criteria. \cite{kourentzes2019another} proposed a heuristic called “forecast islands” to automatically formulate forecast pools and found it beneficial to the accuracy of the combined forecasts. The heuristic proposed by \cite{kourentzes2019another} uses top \textit{q} quantiles for forecast combinations. The cut-off point of how many quantiles is determined automatically rather than arbitrarily. Discarding a set of worst performers is also the most common strategy in the literature on the ``wisdom of crowds'', which is referred to as ``select-crowd'' strategy (see, e.g., \cite{goldstein2014wisdom,mannes2014wisdom,budescu2015identifying}).

\textbf{Robustness.} \cite{lichtendahl2020some} highlighted the importance of robustness in dealing with forecast combination problems. The characteristics of a time series generally change over time, so the pattern detected by a forecasting model in the training set may not continue in the validation and test sets. When dealing with a set of time series, the robustness of a given individual forecast can be assessed using the variance of its accuracy across different series. \cite{lichtendahl2020some} suggested balancing the trade-offs between accuracy and robustness when identifying a subset from the available forecast pool.

\textbf{Diversity.} The performance of forecast combinations also relies on the independent information contained in the individual forecasts, which relates to the diversity of the forecast pool. \cite{mannes2014wisdom} and \cite{thomson2019combining} emphasized that diversity and accuracy are the two crucial factors  in manipulating the quality of the forecast combinations. The benefits of diversity are proved theoretically by \cite{atiya2020does}, who decomposed the mean squared error (MSE) into a bias term and a variance term. \cite{atiya2020does} found that the decrease in the variance of the forecast combination becomes larger as the correlation coefficients among the individual forecasts decrease. In the field of forecast combinations, some scholars have researched to use diversity measures as additional inputs to improve forecast combinations. \cite{kang2022forecast} proposed a diversity-based forecast combination framework using only diversity features. The definition of diversity in \cite{kang2022forecast} is discussed in Section \ref{sec:diversity}.

Until recently, the diversity of the forecast pool has been considered for forecast trimming.
\cite{cang2014combination} designed an optimal subset selection algorithm using mutual information, which measures dependence between individual forecasts. \cite{lichtendahl2020some} eliminated individual forecasts with low accuracy and highly correlated errors. However, considering accuracy and diversity in isolation is questionable, as a poorly performing forecast in the pool may still benefit forecast combinations by injecting diversity into the pool. To solve the problem, \cite{wang2022another} proposed a new algorithm for forecast trimming, taking the robustness and accuracy of the individual forecasts into account, as well as the degree of diversity of the forecast pool, which is discussed in the next section.

\subsection{Forecast trimming}
\label{sec:trimming}
\cite{wang2022another} used the Mean Squared Error for Coherence (MSEC, \cite{thomson2019combining}), also known as Div in \cite{kang2022forecast}, to assess the degree of diversity between individual forecasts. Let $f_{i, h}$ be the $h$th step forecast of the $i$th forecaster in a given forecast pool, where $i=1,2, \ldots, M$ and $h=1,2, \ldots, H$. The MSEC between the $i$th and $j$th methods in the forecast pool is defined as
\begin{equation}
\operatorname{MSEC}_{i, j}=\frac{1}{H} \sum_{h=1}^H\left(f_{i, h}-f_{j, h}\right)^2.
\end{equation}
A value of zero for this measure indicates that the two individuals ($i$ and $j$) have made identical forecasts, and a larger value indicates a higher degree of diversity.

\cite{kang2022forecast} showed that the overall MSE of a weighted combined forecast, $\mathrm{MSE}_{c o m b}$, can be decomposed into component measures corresponding to accuracy (performance) and diversity (coherence), as follows:
\begin{equation}
\operatorname{MSE}_{\text {comb }}  =\sum_{i=1}^M w_i \mathrm{MSE}_i-\sum_{i=1}^{M-1} \sum_{j=2, j>i}^M w_i w_j \mathrm{MSEC}_{i, j}.
\end{equation}
For the process of this decomposition, refer to Equation (\ref{eq:mse}).
Inspired by the decomposition, \cite{wang2022another} proposed a new criterion for forecast trimming to identify an optimal subset for forecast combinations, denoted as Accuracy-Diversity Trade-off (ADT). The ADT criterion is given by
\begin{equation}
\begin{aligned}
\mathrm{ADT} & =\operatorname{AvgMSE}-\kappa \operatorname{AvgMSEC} \\
& =\underbrace{\frac{1}{M} \sum_{i=1}^M \operatorname{MSE}_i}_{\text {mean level of accuracy }}-\kappa \underbrace{ \frac{1}{M^2} \sum_{i=1}^{M-1} \sum_{j=2, j>i}^M \mathrm{MSEC}_{i, j}}_{\text {overall diversity }},
\end{aligned}
\end{equation}
where $\kappa$ is a scale factor and $\kappa \in[0,1]$.

Then a new trimming algorithm based on ADT is proposed, denoted as RAD, which simultaneously addresses robustness, accuracy, and diversity. To apply RAD algorithm, the available in-sample
data needs to be divided into the training set and the validation set. The training set is used to fit statistical or machine-learning models. The forecasts from these fitted models are then evaluated against the validation set, whose length is the same as the required out-of-sample horizon, and then used to identify an optimal subset based on the RAD algorithm.

Firstly, considering an initial forecaster set $\mathbb{S}$, we can apply Tukey’s fences approach to exclude the individuals that lack robustness. Specifically, the individual forecasters with the variance of absolute errors exceeding $Q_3+1.5\left(Q_3-Q_1\right)$ are removed, where $Q_1$ and $Q_3$
are the first and third quantiles of the respective values across individual forecasters from the set $\mathbb{S}$. Then for individual forecaster $i$, calculate the ADT value of the remaining set after removing $i$ from $\mathbb{S}$, find the minimum ADT value and exclude the corresponding forecaster. Repeat the above step until there is a non-significant reduction of the ADT value for $\mathbb{S}$ or until $\mathbb{S}$ contains only two forecasters. The optimal subset identified by RAD approach has been proved to achieve good performance and robustness for both point forecasts and prediction intervals \citep{wang2022another}.

\section{Some Practical Forecasting Issues}
\label{sec:4}

As mentioned above, some scholars used meta-learners to construct feature-based forecast selection/combination methods in recent years, such as using random forecast \citep{talagala2023metalearning}, XGBoost \citep{montero2020fforma,kang2022forecast}, GAMs \citep{wang2022uncertainty}. \cite{montero2020fforma} proposed a feature-based framework for forecast combination named FFORMA, which used meta-learning to model the links between time series features and the out-of-sample performance of individual forecasting models. The following optimization problem is solved to obtain the combination weights:
\begin{equation}
\label{eq:optimization}
 \underset{w}{\arg \min } \sum_{n=1}^N \sum_{i=1}^M w\left(F_n\right)_i \times L_{ni},
\end{equation}
where $F_n$ is the feature vector of the $n$th time series, $N$ is the number of time series, and $M$ is the number of forecasting methods. $L_{ni}$ is the forecast loss of the $i$th method for the $n$th time series.

\cite{montero2020fforma}'s method placed second in the M4 competition
and inspired a series of expanded studies. \cite{li2020forecasting} introduced an automated approach to extract time series features based on time series imaging, as discussed in \ref{sec:imaging}. \cite{kang2022forecast} developed FFORMA by using diversity to replace time series features in Equation \ref{eq:optimization}. It can reduce the computational complexity and compute automatically without the need for manual choice of an appropriate set of features. See \ref{sec:diversity} for more details. \cite{montero2020fforma}, \cite{kang2022forecast} and \cite{li2020forecasting}'s work are all based on M4 competition dataset, containing 100,000 time series with different seasonal periods from different domains such as demographics, finance, and industries. In addition, this chapter discusses other applications of the aforementioned feature-based methods in the following complex issues.

Demand forecasting is the basis for most planning and control activities in any organization.  In particular, demand may appear sporadically, with no demand in some periods, leading
to an intermittent appearance. Intermittent demand items are prevalent in many industries, including the automotive, IT, and electronics sectors. Their inventory implications are dramatic and forecasting their requirements is challenging. \cite{li2022feature} examined the empirical outcomes of some existing forecast combination methods and proposed a generalized
feature-based framework for intermittent demand forecasting, which is discussed in Section \ref{sec:Intermittent}.


In recent years, probabilistic forecasts have received increasing attention. For example, the recent M4 and the M5 uncertainty competitions \citep{makridakis2022m5} encouraged participants to provide probabilistic forecasts of different types as well as point forecasts. Probabilistic forecasts are appealing for enabling optimal decision-making with a better understanding of uncertainties and risks. Interval forecasts form a special case of probabilistic forecasts, which are evidently easier to implement. \cite{wang2022uncertainty}  investigated how time series features affect the performances of prediction intervals from different methods. The proposed feature-based interval forecasting framework is discussed in Section \ref{sec:Uncertainty}.

\subsection{Intermittent demand}
\label{sec:Intermittent}
What makes intermittent demand challenging to forecast is that there are two sources of uncertainty: the sporadic demand occurrence, and the demand arrival timing. Despite that intermittent demand forecasting has obtained some research achievements in recent decades, limited attention has been given to combination schemes for intermittent demand forecasting. \cite{li2022feature} provided a discussion and comparison of forecast combination methods in the context of intermittent demand forecasting and developed a feature-based
combination framework for intermittent demand. The proposed framework has been shown to improve the accuracy of point and quantile forecasts.

\cite{li2022feature} defined a broad pool for intermittent demand forecasting. The pool includes traditional forecasting models and intermittent demand forecasting methods to ensure diversity in the pool. The authors built an XGBoost model to learn the relationship between features and combination weights. Time series features and the diversity of the pool of methods are all valid inputs for the forecast combination model. \cite{li2022feature} considered nine explainable time series features, which imply the intermittency, volatility, regularity and obsolescence of intermittent demand, and applied the scaled diversity defined in \cite{kang2022forecast} for intermittent demand forecasting.

In the literature, some scholars argue that traditional accuracy
measures based on absolute errors are unsuitable for intermittent
demand, because a flat zero forecast is frequently ``best'' for these measures when the demand is highly intermittent \citep{kolassa2016evaluating}. The root mean squared scaled error (RMSSE) is used to measure the performance of point forecasts for intermittent demand in \cite{li2022feature}'s work, which can be obtained as:
\begin{equation}
R M S S E=\sqrt{\frac{1}{H} \frac{\sum_{h=1}^H\left(y_{T+h}-\hat{y}_{T+h}\right)^2}{\frac{1}{T-1} \sum_{t=2}^T\left(y_t-y_{t-1}\right)^2}},
\end{equation}
where $H$ is the forecasting horizon. $\hat{y}_{T+h}$ is the $h$th step forecast generated from a series of observed values $\left\{y_t, t=1,2, \cdots, T\right\}$, and $y_{T+h}$ is the true value. RMSSE focuses on the expectation, which is consistent with the candidate methods in the forecasting pool.

The feature-based framework for intermittent demand is an extension of Figure~\ref{fig:framework} by customizing a forecasting pool, a loss function based on RMSSE and time series features. The procedures of the aforementioned framework are formulated into the \pkg{fide} R package available at \url{https://github.com/lily940703/fide}.

\subsection{Uncertainty estimation}
\label{sec:Uncertainty}

The literature on the uncertainty estimation of feature-based time series forecasts is limited compared to point forecasting. The M4 forecasting competition \citep{makridakis2020m4} encouraged participants to provide point forecasts and prediction intervals (PIs). Among the submissions, \cite{montero2020fforma} computed the point forecasts using FFORMA and obtained the PIs by using a simple equally weighted combination of the 95\% bounds of naïve, theta and seasonal naïve methods. This approach ranked second in the M4 competition but did not consider any time series features when calculating the interval forecasts. \cite{wang2022uncertainty} applied meta-learners to explore how time series features affect the uncertainty estimation of forecasts, which is measured by PIs.  GAMs \citep{hastie2017generalized} are applied to depict the relationship between time series features and interval forecasting accuracies, making interval forecasts interpretable for time series features.

\cite{wang2022uncertainty} proposed a feature-based framework for time series interval forecasting. The reference dataset used for training the algorithm is generated by GRATIS approach \citep{kang2020gratis}. The authors used a set of 42 features which are the same as the features in \cite{montero2020fforma}.  The mean scaled interval score (MSIS) is used to measure the accuracy of PIs. The calculation of MSIS can be stated as follows:
\begin{equation}
MSIS=\frac{1}{h} \frac{\sum_{t=n+1}^{n+h}\left(U_t-L_t\right)+\frac{2}{\alpha}\left(L_t-Y_t\right) \mathbb{1}\left\{Y_t<L_t\right\}+\frac{2}{\alpha}\left(Y_t-U_t\right) \mathbb{1}\left\{Y_t>U_t\right\}}{\frac{1}{n-m} \sum_{t=m+1}^n\left|Y_t-Y_{t-m}\right|},
\end{equation}
where $Y_t$ are the true values of the future data, $\left[L_t, U_t\right]$
are the generated PIs, $h$ is the forecasting horizon, $n$ is the length of the historical data, and $m$ is the time interval symbolizing the length of the time series periodicity (e.g., m takes the values of 1, 4, and 12 for
yearly, quarterly, and monthly data, respectively). $\mathbb{1}$ is
the indicator function, which returns 1 when the condition
is true and otherwise returns 0.

Based on the error measure of  MSIS, \cite{wang2022uncertainty}  established the relationship between the mean values of log(MSIS) and the extracted features by using GAMs. The authors also selected a subset of appropriate methods for each time series tailored to their features from the method pool using an optimal threshold ratio searching algorithm. An R package \pkg{fuma} implements the aforementioned framework, which is available at \url{https://github.com/xqnwang/fuma}.

\section{Conclusion}

In the era of big data, the information from the whole dataset is often used to improve forecasts.  With the development of AI, a series of feature-based methods have sprung up. Scholars use meta-learning to describe the relationship between features and forecast model selection/combination. There is a general framework for feature-based methods, which contains the model training and forecasting phases.  Firstly, one needs to determine a set of diverse reference data to train the meta-learner. Generating sythetic time series  with \pkg{gratis} provides an efficient way to generate time series with diverse and controllable characteristics. Secondly, we select a set of features used to describe the time series characteristics for forecasting. In addition to selecting appropriate features manually based on expert knowledge, the automation of feature extraction has gotten more attention in recent years. Advanced methods include time series imaging, calculation of forecast diversity and automatic feature selection procedures. Thirdly, deciding on a forecast pool is necessary, especially when there is a plethora of alternative methods. Accuracy, robustness, and diversity of the individual forecasts are the most critical factors that affect the pool's forecasting performance.  A recently proposed algorithm named RAD can address the three factors simultaneously and identify an optimal subset from the initial forecast pool.

Based on the above reference set, selected features and trimmed forecast pool, a meta-learner can be trained based on features and forecast errors and obtain a forecast selection/combination model. The forecasting phase calculates the features for the new time series and gets the optimal method or combination weights through the pre-trained model. The feature-based framework is feasible with adjustable forecast pools, features, error measures and meta-learners, which has been extended to practical issues, such as intermittent demand forecasting and uncertainty estimation.

However, one may also notice the methods in this chapter are mostly for point forecasting with determinist features. Probabilistic time series forecasting that takes feature uncertainty into consideration has seldom been studied in the literature. More study is required for interpreting features and the associated weights. Solely using time series features to forecast is also a promising direction when data privacy is a concern in the big data era.

\section*{Acknowledgments}
Feng Li's research was supported by the National Social Science Fund of China (22BTJ028).

\bibliographystyle{spbasic}
\bibliography{ref.bib}

\end{document}